%% file: main.tex
\documentclass{article}

\usepackage{microtype}
\usepackage{marvosym}
\usepackage{graphicx}
\usepackage{subcaption}
\usepackage{booktabs}
\usepackage{multirow}
\usepackage{enumitem}
\usepackage{amssymb}
\usepackage{hyperref}

\usepackage[preprint]{icml2026}
\usepackage{amsmath}
\usepackage{amssymb}
\usepackage{mathtools}
\usepackage[capitalize,noabbrev]{cleveref}

\icmltitlerunning{FORCE: Efficient VLA Reinforcement Fine-Tuning}

\begin{document}

\twocolumn[
  \icmltitle{FORCE: Efficient VLA Reinforcement Fine-Tuning via Value-Calibrated Warm-up and Self-Distillation}

 \icmlsetsymbol{equal}{*}
  \newcommand{\leadsymbol}{\raisebox{0.15ex}{\scalebox{0.72}{$\dagger$}}}
\icmlsetsymbol{lead}{\hspace{-0.08em}\leadsymbol\hspace{-0.03em}}
    % \begin{icmlauthorlist}
    %     \icmlauthor{Yunfan Lou}{yyy}
    %     \icmlauthor{Xiaowei Chi}{yyy}
    %     \icmlauthor{Xiaojie Zhang}{yyy}
    %     \icmlauthor{Zezhong Qian}{yyy}
    %     \icmlauthor{Chengxuan Li}{yyy}
    %     \icmlauthor{Rongyu Zhang}{yyy}
    %     \icmlauthor{Yaoxu Lyu}{yyy}
    %     \icmlauthor{Guoyu Song}{yyy}
    %     \icmlauthor{Chuyao Fu}{yyy}
    %     \icmlauthor{Haoxuan Xu}{yyy}
    %     \icmlauthor{Pengwei Wang}{yyy}
    %     \icmlauthor{Shanghang Zhang}{yyy}
    % \end{icmlauthorlist}

    \begin{icmlauthorlist}
        \icmlauthor{Shuyi Zhang}{equal,casia,baai}
        \icmlauthor{Yunfan Lou}{equal,baai}
        \icmlauthor{Hongyang Cheng}{equal,baai}
        \icmlauthor{Yichen Guo}{pku_skl}
        \icmlauthor{Chuyao Fu}{baai}
        \icmlauthor{Yaoxu Lyu}{baai,pku_skl}
        \icmlauthor{Xiaojie Zhang}{baai}
        \icmlauthor{Haoran Li}{casia}
        \icmlauthor{Pengwei Wang}{baai}
        \icmlauthor{Zhongyuan Wang}{baai}
        \icmlauthor{Shanghang Zhang\textsuperscript{\Letter}}{baai,pku_skl}
    \end{icmlauthorlist}

    \icmlaffiliation{casia}{Institute of Automation, Chinese Academy of Sciences, Beijing, China}
    \icmlaffiliation{baai}{Beijing Academy of Artificial Intelligence, Beijing, China}
    \icmlaffiliation{pku_skl}{State Key Laboratory of Multimedia Information Processing, School of Computer Science, Peking University, Beijing, China}

  % \begin{icmlauthorlist}
  %   \icmlauthor{Firstname1 Lastname1}{equal,yyy}
  %   \icmlauthor{Firstname2 Lastname2}{equal,yyy,comp}
  %   \icmlauthor{Firstname3 Lastname3}{comp}
  %   \icmlauthor{Firstname4 Lastname4}{sch}
  %   \icmlauthor{Firstname5 Lastname5}{yyy}
  %   \icmlauthor{Firstname6 Lastname6}{sch,yyy,comp}
  %   \icmlauthor{Firstname7 Lastname7}{comp}
  %   %\icmlauthor{}{sch}
  %   \icmlauthor{Firstname8 Lastname8}{sch}
  %   \icmlauthor{Firstname8 Lastname8}{yyy,comp}
  %   %\icmlauthor{}{sch}
  %   %\icmlauthor{}{sch}
  % \end{icmlauthorlist}


  \icmlcorrespondingauthor{Shanghang Zhang}{shanghang@pku.edu.cn}

  \icmlkeywords{Machine Learning, ICML}

  \vskip 0.3in
]



% \printAffiliationsAndNotice{\icmlEqualContribution}  % no special notice (required even if empty)
\printAffiliationsAndNotice{\icmlEqualContribution\ \textsuperscript{\Letter} Corresponding author}

\input{sec/0_abstract}

\input{sec/1_intro}
\input{sec/2_related_work}

\input{sec/3_method}

\input{sec/4_experiment}
\input{sec/5_conclusion}

\bibliography{main}
\bibliographystyle{icml2026}

%%%%%%%%%%%%%%%%%%%%%%%%%%%%%%%%%%%%%%%%%%%%%%%%%%%%%%%%%%%%%%%%%%%%%%%%%%%%%%%
% APPENDIX (included in the same PDF for submission)
%%%%%%%%%%%%%%%%%%%%%%%%%%%%%%%%%%%%%%%%%%%%%%%%%%%%%%%%%%%%%%%%%%%%%%%%%%%%%%%
\newpage
\appendix
\onecolumn
\input{sec/X_suppl}

\end{document}

%% file: sec/0_abstract.tex
% 发展现状1 + (问题1) + we present 1 + method 3-4 + 优点 1-2 （一半篇幅）+ we evaluate on 1 + result 1  + conclusion 1 + website 1

\begin{abstract}

 Vision-Language-Action (VLA) models are often constrained by the 'imitation ceiling' imposed by sub-optimal data. While Reinforcement Learning (RL) fine-tuning can surpass this limit, it is notoriously sample inefficient. This challenge arises from two core issues: (1) catastrophic initial unlearning due to an unstable Q-function and (2) inefficient policy updates caused by low-quality exploration data, often forcing a reliance on costly human interventions. We introduce \textbf{FORCE}, a 3-stage framework that stabilizes fine-tuning by tackling both issues. FORCE first incorporates a \textit{Value-Calibrated Warm-Up} phase, utilizing on-policy rollouts to mitigate the distributional shift of the Q-function. Subsequently, during the online stage, this calibrated Q-function acts as a filter for both the policy's own action proposals and expert data, ensuring only high-value actions are used for the policy update. We evaluate FORCE on various simulation and real-world tasks, and the result shows that FORCE achieves a 79\% absolute improvement in success rates and outperform prior RL methods by 10\%, while accelerating training by 32.5\%. Critically, it mitigates the common success rate drop and achieves this robust performance without human intervention, marking a significant step towards deploying capable and autonomous robotic agents.

\end{abstract}

%% file: sec/1_intro.tex
\section{Introduction}
\label{sec:intro}

\begin{figure*}[ht]
  \centering
  \includegraphics[width=1.0\textwidth]{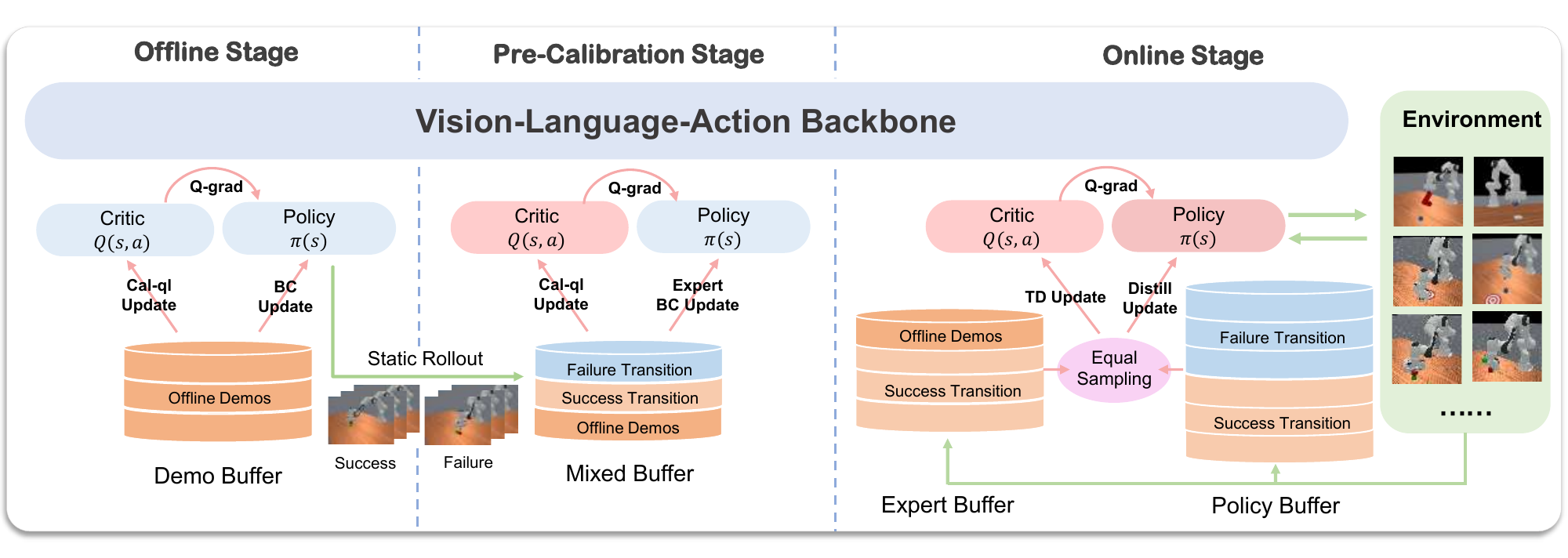}
  \caption{\textbf{Overview of the FORCE framework.} Our method employs a three-stage reinforcement fine-tuning pipeline that progressively calibrates value estimation and stabilizes policy improvement: (1) offline Cal-QL pretraining to obtain a conservative and well-grounded critic, (2) mixed-rollout value pre-calibration to bridge offline and online distributions and mitigate O2O drift, and (3) online fine-tuning with balanced expert/policy replay coupled with VGPD, enabling safe and sample-efficient policy refinement beyond the imitation ceiling.}
  \label{fig:main_method}
  \vspace{-1em}
\end{figure*}

Vision-Language-Action models have demonstrated remarkable potential in executing general-purpose robotic tasks \cite{kim2024openvla, zitkovich2023rt-2, ghosh2024octo}. These models are typically pre-trained through Imitation Learning (IL), enabling robots to understand complex instructions and map them to physical actions \cite{blackP0VisionLanguageActionFlow, blackP05VisionLanguageActionModel}. However, pure imitation learning (such as Supervised Fine-Tuning, SFT) faces a fundamental bottleneck: the policy's performance is constrained by the quality of the demonstration data, known as the "imitation ceiling" \cite{ross2011reduction}. This is not merely an empirical observation but a fundamental statistical limitation, as imitation learning suffers from error compounding that scales quadratically with task horizon \cite{rajaraman2020limits}. Since human demonstration data inevitably contains sub-optimal and even inconsistent actions \cite{hester2018deep, chen2021decision}, which standard IL struggles to learn from without additional information like trajectory rankings \cite{guhur2023instruction}, the performance ceiling of VLA models is firmly capped.

% 1.29 这个地方可以先讲RL再讲VLA，这样会让vla成为rl的setting，重点变到rl
% Reinforcement learning (RL) is a principled way to improve robotic policies beyond what is present in a fixed dataset, by leveraging trial-and-error interaction and long-horizon credit assignment \cite{haarnoja2018sac, vecerik2017ddpgfd, kalashnikov2018qtopt}. 
% In practice, however, deploying RL on physical robots remains notoriously sample-inefficient and safety-critical, making stable and data-efficient fine-tuning a central bottleneck for real-world robotics.

% Modern Vision-Language-Action (VLA) models provide a strong starting point for this paradigm, by mapping natural language instructions and visual observations to continuous robot actions \cite{kim2024openvla, zitkovich2023rt-2, ghosh2024octo, blackP0VisionLanguageActionFlow, blackP05VisionLanguageActionModel}. 
% While imitation learning offers a convenient initialization, it is fundamentally bounded by the quality of the collected behaviors (the ``imitation ceiling'') \cite{ross2011reduction, rajaraman2020limits}. 
% In real robot datasets, demonstrations are often sub-optimal or inconsistent \cite{hester2018deep, chen2021decision}, which limits pure imitation even when using large-capacity VLA policies.

Reinforcement learning offers a clear path for VLA models to surpass the limitations of demonstration data through online interaction with the environment, thereby breaking the "imitation ceiling" \cite{vecerik2017ddpgfd, kalashnikov2018qtopt}. Indeed, recent works have shown that even a small amount of online fine-tuning can dramatically improve the performance of large-scale, pre-trained IL policies \cite{li2025simplevla-rl, hu2025flare, lei2025rl-100}. Despite this promise, applying RL directly to the online fine-tuning of VLA models faces severe challenges. The most prominent issue is that RL fine-tuning is notoriously sample inefficient in real-world physical interactions \cite{haarnoja2018sac}. This sample inefficiency remains the primary bottleneck, even as recent reinforced fine-tuning (RFT) frameworks have shown impressive gains with just minutes of online interaction \cite{chen2025conrft, openvla2024oft}.
% 1.29 这里也是把rl4vla改为vla4rl
% RL therefore offers a clear path to surpass demonstration limitations by improving policies through interaction, potentially discovering behaviors absent from the dataset \cite{vecerik2017ddpgfd, kalashnikov2018qtopt}. 
% Recent work has further shown that starting from a strong pre-trained policy (e.g., VLA) can substantially reduce unnecessary exploration and enable meaningful gains with limited online data \cite{li2025simplevla-rl, hu2025flare, lei2025rl-100}. 
% Despite this promise, sample efficiency remains the primary bottleneck in real-world RL: even brief online interaction is costly, slow, and safety-constrained on physical hardware \cite{haarnoja2018sac}. 
% This bottleneck persists even for recent reinforced fine-tuning (RFT) pipelines that report rapid gains under carefully designed training recipes \cite{chen2025conrft, openvla2024oft}.

The root of this inefficiency can be traced to two core difficulties in offline-to-online RL on real robots. (1) \textbf{Catastrophic 'Initial Unlearning'}: When transitioning from offline pre-training (which often uses conservative value estimates) to online fine-tuning, policies suffer from a well-documented performance collapse \cite{nakamoto2023calql,nair2020awac}. This "initial unlearning"  is caused by a Q-value scale mismatch, where the highly underestimated offline Q-function is "deceived" by new online data, leading to a catastrophic adjustment period and negating the benefits of pre-training. (2) \textbf{Inefficient Policy Updates}: The policy update process is severely hindered by low-quality exploratory data (i.e., the policy generates a large number of useless actions when exploring unfamiliar regions). To address these problems and ensure safe exploration, existing methods \cite{luo2025precise, chen2025conrft, kelly2019hg} are often forced to rely on costly and difficult-to-scale Human-in-the-Loop (HiL) interventions \cite{luo2025precise}, which treat human attention as a scarce resource.

To address these challenges, we introduce FORCE, an intervention-free framework that bridges the gap between static datasets and dynamic physical interaction. To mitigate catastrophic unlearning, FORCE incorporates a \textit{distributional warm-up} phase that calibrates the value function on the policy's visitation distribution prior to any actor updates. This aligns the critic's support with the agent's current behavior, ensuring stable value estimates during the offline-to-online transition. Subsequently, the framework employs a value-guided self-distillation mechanism to govern online fine-tuning. Acting as a dynamic filter, this module projects the policy onto a target distribution derived strictly from high-value transitions, effectively eliminating exploration noise and ensuring monotonic improvement without human supervision.

% TODO: 这个地方明确一下前面提到的2个问题是如何解决的
% The core mechanism of FORCE is shown in Fig.~\ref{fig:main_method} as follows: First, we introduce a crucial "transition stage" between traditional offline pre-training and online fine-tuning. During this stage, we perform "value pre-calibration," with the objective of obtaining a stable and reliable Q-function \textit{before} engaging in costly online interaction with the real environment. Subsequently, during the online fine-tuning stage, this calibrated Q-function acts as a "filter." It not only evaluates and filters the actions generated by the policy's own exploration but also re-evaluates the expert data from the offline dataset. This dual-filtering approach builds on emerging ideas of value-based data filtering \cite{mark2024policy} but applies them symmetrically to both online and offline data streams. In this way, FORCE ensures that only high-value actions (whether from exploration or prior data) are used for the policy update, thereby guaranteeing both the stability and efficiency of the update.

We evaluated FORCE on several challenging robotic manipulation tasks both in simulation and the real-world. The experimental results demonstrate that FORCE achieves significant performance gains, reaching nearly 100\% success rates on several tasks, all without human intervention. Our main contributions are summarized as follows:

% \begin{enumerate}
%     \item \textbf{FORCE Framework:} A principled three-stage pipeline for intervention-free offline-to-online RL fine-tuning starting from pre-trained models (e.g., VLA).
%     \item \textbf{Distributional Warm-up:} A mechanism to mitigate cold-start covariate shift by aligning Q-function support with policy visitation prior to fine-tuning.
%     \item \textbf{Value-Guided Policy Distillation:} A theoretically grounded regularized policy improvement operator that filters high-variance exploration to accelerate convergence.
%     \item \textbf{SOTA Performance and Efficient Adaptation:} Empirical validation on physical robots showing that FORCE achieves near-perfect success rates on contact-rich tasks with significantly improved sample efficiency compared to baselines.
% \end{enumerate}
\begin{enumerate}[itemsep=2pt, topsep=4pt, parsep=1pt]
    \item \textbf{FORCE Framework:} A principled three-stage pipeline for intervention-free offline-to-online RL fine-tuning starting from pre-trained models (e.g., VLA).
    \item \textbf{Distributional Warm-up:} A mechanism to mitigate cold-start covariate shift by aligning Q-function support with policy visitation prior to fine-tuning.
    \item \textbf{Value-Guided Policy Distillation:} A theoretically grounded regularized policy improvement operator that filters high-variance exploration to accelerate convergence.
    \item \textbf{SOTA Performance and Efficient Adaptation:} Empirical validation on physical robots showing that FORCE achieves near-perfect success rates on contact-rich tasks with significantly improved sample efficiency compared to baselines.
\end{enumerate}

% \begin{enumerate}
%     \item We propose FORCE, a novel three-stage VLA fine-tuning framework that stably improves policy performance without human intervention.
%     \item We introduce a \textbf{value-calibrated warm-up} stage, specifically designed to stabilize the Q-function before online learning begins, addressing the severe noise problem in traditional RL fine-tuning value estimates.
%     \item We designed a \textbf{value-guided self-distillation mechanism} based on the calibrated Q-function, which can simultaneously purify policy exploration data and existing expert data, significantly enhancing the efficiency of policy updates.
%     \item We validated the effectiveness of FORCE in real-world robotic tasks, demonstrating its ability to achieve SOTA performance levels without the need for human intervention.
% \end{enumerate}

% 1.29 这里一样
% We propose FORCE, a novel, intervention-free three-stage offline-to-online RL framework that stably improves real-robot policies starting from pre-trained models (e.g., VLA).

%% file: sec/2_related_work.tex
\section{Related Work}
\label{sec:related}

\noindent\textbf{Vision-Language-Action Models} Vision-Language-Action (VLA) Models have recently enabled robots to perceive, reason, and act using multimodal inputs~\cite{brohan2022rt, zitkovich2023rt-2}. Early architectures treated manipulation as a discrete token generation problem, like OpenVLA \cite{kim2024openvla}, often resulting in coarse control resolution. This limitation motivated models to utilize flow matching or diffusion processes to chunk continuous motions into high-dimensional latent flows, such as Octo \cite{ghosh2024octo}, Rdt-1b \cite{liu2024rdt} and $\pi$-series models~\cite{blackP0VisionLanguageActionFlow, blackP05VisionLanguageActionModel, intelligence2025pi}. However, these models predominantly rely on supervised imitation learning from static datasets, which prevents them from correcting execution errors or adapting to dynamic environments. In our work, we focus on fine-tuning high-capacity VLA policies using RL to further enhance their performance and generalization.

% \noindent\textbf{Vision-Language-Action Models} Vision-Language-Action (VLA) Models have recently enabled robots to perceive, reason, and act using multimodal inputs. A number of VLA architectures have been proposed to unify vision, language, and control, ranging from discrete action token models to continuous flow-based policies. For example, OpenVLA (and similar autoregressive models) treats low-level actions as part of the model’s text-based vocabulary, allowing language-conditioned action generation but at a coarse resolution. This limitation motivated the $\pi$-series models ($\pi_0$ , $\pi_{0.5}$ \cite{blackP05VisionLanguageActionModel}, \cite{blackP0VisionLanguageActionFlow}) and others that use flow-matching or diffusion policies to produce continuous action trajectories for more fine-grained and dexterous manipulation. These flow-based VLA models chunk continuous motions into high-dimensional latent flows, achieving more expressive control than token discretization. In our work, we focus on fine-tuning such high-capacity VLA policies using online RL to further enhance their performance and generalization through interaction.

\noindent\textbf{RL Finetuning for VLA Models and Real-world RL} Many works recently have explored reinforcement learning algorithm to improve the generalization ability of VLA models. Approaches building on PPO \citep{schulman2017ppo} or GRPO \cite{shao2024deepseekmath}s with transformer backbones remain largely restricted to simulation~\cite{li2025simplevla-rl, lu2025vla, liu2025can}, as their on-policy nature suffers from poor sample efficiency that prohibits physical deployment. 
Moreover, these standard formulations are structurally limited to autoregressive backbones, failing to accommodate continuous policies required for dexterous manipulation. Real-world RL training strategies for VLA models generally fall into two paradigms: (1) \textbf{generalized behavioral cloning}~\cite{nair2020awac, peng2019awr}, exemplified by methods such as PA-RL\cite{mark2024policy}, which utilizes value preference feedback to steer the policy towards high-quality behaviors while maintaining the stability of supervised imitation; and (2) \textbf{direct Q-driven optimization augmented with behavioral cloning regularization} \cite{fujimoto2021minimalist}, like ConRFT \cite{chen2025conrft}, is to synergize the stability of behavioral cloning (BC) with the optimality-seeking drive of Q-learning via a joint optimization objective. Our framework, FORCE, adopts this hybrid objective but additionally employs a novel value-guided self-distillation mechanism in the online phase to mitigate instability and maximize sample efficiency.

\noindent\textbf{Offline-to-Online Shift}  \indent Training robotic policies from scratch is notoriously sample-inefficient and often demands extensive, infeasible human-in-the-loop intervention\cite{luo2025precise,chen2025conrft}. The offline-to-online fine-tuning paradigm has emerged as a compelling alternative\cite{yu2023actor, zheng2023adaptive, song2022hybrid, kumar2020conservative}, yet it is plagued by the "initial unlearning" phenomenon \cite{kumar2020cql}. This critical instability arises when the policy, pre-trained on static data, first encounters the online environment. To bridge this gap, RL100\cite{lei2025rl-100} interleaves imitation re-learning in the offline setting. Inspired by this, our framework introduces a novel value pre-calibration stage. By strategically augmenting the offline phase with diverse, sub-optimal trajectories, we effectively pre-stabilize the Q-function's distribution before online fine-tuning begins. This pre-calibration acts as a crucial transition step, significantly mitigating initial unlearning and enabling a much smoother and more stable handoff from offline to online learning.

%% file: sec/3_method.tex
\section{Method}
\label{sec:method}
We articulate a three-stage training paradigm spanning offline reinforcement learning and an offline-to-online bridge phase followed by online fine-tuning in interactive environments. To further bolster sample efficiency during online learning, we introduce a value-guided policy self-distillation (VGPD) method. In the following section, we provide a detailed account of the training pipeline and the VGPD module.

\subsection{Problem Setup}

We address the problem of fine-tuning pretrained VLA models for complex manipulation tasks. While SFT on offline expert demonstrations is standard, it suffers from the \textit{distributional shift} problem: the learned policy $\pi$ inevitably visits states outside the support of the expert data $\mathcal{D}_E$, leading to compounding errors. To mitigate this, we formulate the fine-tuning process as a Markov Decision Process (MDP) $\mathcal{M}=(\mathcal{S},\mathcal{A},P,R,\gamma)$. The objective is to learn a policy $\pi$ that maximizes the expected discounted return $J(\pi) = \mathbb{E}_{\pi}[\sum_{t=0}^\infty \gamma^t R(s_t,a_t)]$. We define the state-action value function $Q^\pi(s,a) = \mathbb{E}_{\pi}[\sum_{t=0}^\infty \gamma^t R(s_t,a_t) | s_0=s, a_0=a]$.

Our framework addresses two fundamental challenges in VLA fine-tuning: (1) \textbf{Support Mismatch}, where the Q-function is undefined on the current policy's visitation distribution $d^{\pi}$ during the initial offline-to-online transition; and (2) \textbf{High-Variance Exploration}, where standard online policy gradient methods fail due to the sparsity of rewards and the high dimensionality of the VLA action space.

\subsection{Training pipeline}

\noindent\textbf{\textit{S}1. Offline Reinforcement Fine-Tuning with Expert Demonstrations}.

To circumvent the prohibitive exploration costs associated with learning from scratch, we initialize the policy $\pi_{\text{pre}}$ via offline RL on expert demonstrations. We adopt Calibrated Q-Learning (Cal-QL) to learn a conservative value function from the static dataset $\mathcal{D}_{\text{E}}$. Cal-QL minimizes the temporal-difference (TD) error augmented with a calibration regularizer that constrains value estimates based on the behavior policy $\mu_{\mathcal{D}}$. This constraint prevents the overestimation of out-of-distribution (OOD) actions, a critical requirement for stable offline-to-online transfer. The objective for the critic parameter $\theta$ is:
\begin{equation}
\begin{aligned}
    \mathcal{L}_{Q}^{S_1}(\theta)=&\mathbb{E}_{(s,a,s')\sim\mathcal{D_{\text{E}}}}[(Q_{\theta}(s,a)-\mathcal{B}^{\pi}\overline{Q}_{\overline{\theta}}(s,a))^2] \\
    +&\alpha(\mathbb{E}_{s\sim\mathcal{D_{\text{E}}},a\sim\pi(\cdot|s)}[\max(Q_{\theta}(s,a),Q^{\mu}(s,a))]\\
    -&\mathbb{E}_{s,a\sim\mathcal{D_{\text{E}}}}[Q_{\theta}(s,a)])\\
\end{aligned}
    \label{eq:calql_offline}
\end{equation}
Here, the first term represents the standard Bellman error, while the second term imposes the calibration constraint controlled by hyperparameter $\alpha$.

While Cal-QL effectively regularizes the value landscape, optimizing the policy solely against a limited dataset can lead to optimization instability. We therefore employ a behavior-regularized policy optimization objective. This combines a Behavior Cloning (BC) term, which anchors the policy to the demonstrated manifold, with a Q-guided policy gradient term that drives performance maximization. The offline actor objective is defined as:
\begin{equation}
\begin{aligned}
    \mathcal{L}_{\pi}(\phi)^{S_1}
    = \eta\,\mathcal{L}_{\pi}^{BC,S_1}(\phi) + \lambda\,\mathcal{L}_{\pi}^{Q,S_1}(\phi)
\end{aligned}
    \label{eq:offline_actor_loss}
\end{equation}
where
\begin{equation}
\begin{aligned}
    \mathcal{L}_{\pi}^{BC,S_1}(\phi)
    = \mathbb{E}_{(s,a)\sim \mathcal{D_{\text{E}}}}\!\big[-\log \pi_\phi(a\mid s)\big]
\end{aligned}
    \label{eq:actor_bc_loss}
\end{equation}
\begin{equation}
\begin{aligned}
    \mathcal{L}_{\pi}^{Q,S_1}(\phi)
    = -\,\mathbb{E}_{s\sim \mathcal{D_{\text{E}}},\,a\sim \pi_\phi(\cdot\mid s)}\!\big[Q_\theta(s,a)\big]
\end{aligned}
    \label{eq:actor_q_loss}
\end{equation}
Here, $\eta$ and $\lambda$ modulate the trade-off between imitation and maximization. In contemporary VLA models, the action head is often implemented as a multi-step diffusion or flow-matching denoiser, which substantially increases computational cost due to iterative denoising. Distilling a diffusion/flow policy into a one-step policy has been shown to be an effective alternative: it largely preserves the expressiveness of the original policy while avoiding iterative denoising and BPTT problem. In this work, we adopt a consistency policy as the actor network, which maintains the original performance while enabling fast inference and more stable Q-gradient propagation.

\noindent{\textbf{\textit{S}2. Mitigating Distributional Shift via On-Policy Warm-up}}.

\begin{figure}[!ht]
\centering
\vspace{-3mm}
\includegraphics[width=0.9\linewidth]{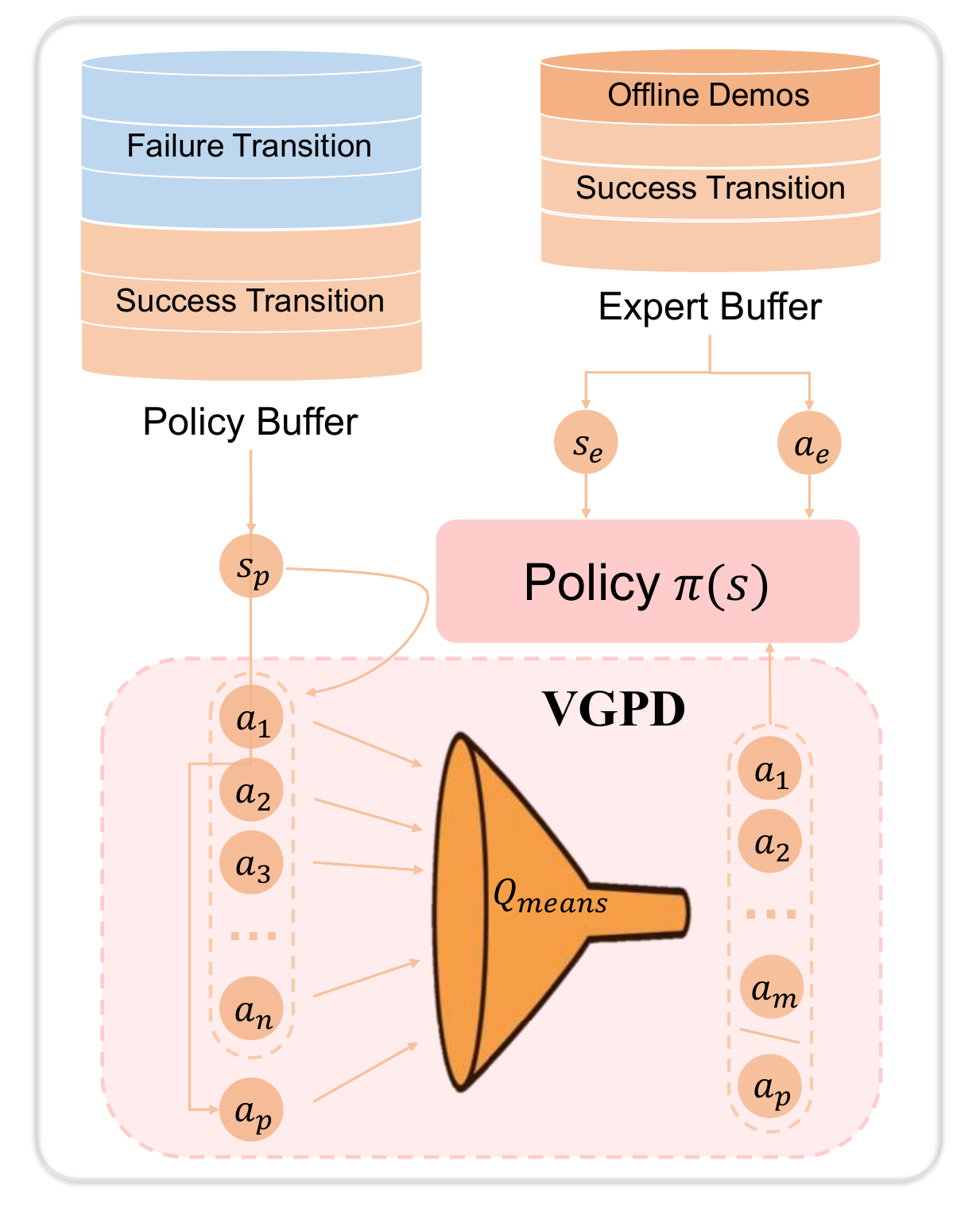}
\vspace{-2mm}
\caption{\textbf{VGPD module in the online phase.}
  VGPD serves as a regularized policy improvement mechanism. We maintain an expert buffer and a policy buffer. For states sampled from the policy buffer, we compute a dynamic value baseline $V_{\text{ref}}(s)$ (approximated by $Q_{\text{mean}}$). The policy is updated via filtered importance sampling, distilling only from actions that show positive advantage over this baseline.}
\label{fig:distll_method}
\vspace{-3mm}
\end{figure}

A critical challenge in offline-to-online (O2O) RL is the initial performance collapse caused by \textit{Covariate Shift}. This phenomenon arises from the distributional mismatch between the offline dataset $\mathcal{D}_{\text{offline}}$, sampled from behavior policy $\pi_\beta$, and the visitation distribution $d^{\pi_\phi}$ of the current policy. Standard offline algorithms constrain $\pi_\phi$ close to $\pi_\beta$, but strictly enforcing this constraint inhibits necessary exploration and correction.

In Stage 2, we introduce a \textit{Distributional Warm-up}. We collect a small batch of on-policy trajectories $\mathcal{D}_{\text{warm}} = \{ (s, a, r, s') \sim \pi_\phi \}$ and merge them with the offline data: $\mathcal{D}_{\text{mix}} = \mathcal{D}_{\text{offline}} \cup \mathcal{D}_{\text{warm}}$. Crucially, we apply the conservative value constraint (Eq. \ref{eq:calql_offline}) on this mixed dataset. This explicitly expands the valid support of the calibrated Q-function from $\text{supp}(\pi_\beta)$ to $\text{supp}(\pi_\beta) \cup \text{supp}(\pi_\phi)$. By doing so, we ensure that Q-estimates remain well-defined and lower-bounded on the manifold of states the policy is actually visiting, addressing the support mismatch issue before full online adaptation.

Consequently, the critic’s ranking of $(s,a)$ pairs aligns with the on-policy distribution, yielding a robust value function that prevents catastrophic unlearning. The actor is trained with an asymmetric objective:
\begin{equation}
    \begin{aligned}
    \mathcal{L}_{\pi}^{S_2}(\phi)
    = \eta\,\mathcal{L}_{\pi}^{BC,S_2}(\phi) + \lambda\,\mathcal{L}_{\pi}^{Q,S_2}(\phi)
    \end{aligned}
    \label{eq:actor_s2_obj}
\end{equation}
where
\begin{equation}
    \begin{aligned}
    \mathcal{L}_{\pi}^{BC,S_2}(\phi)
    = \frac{1}{\rho}\,\mathbb{E}_{(s,a,y)\sim \mathcal{D}_{\text{mix}}}\!\big[\mathbf{1}\{y{=}1\}\cdot(-\log \pi_\phi(a\mid s))\big]
    \end{aligned}
    \label{eq:actor_bc_succ}
\end{equation}
\begin{equation}
    \begin{aligned}
    \mathcal{L}_{\pi}^{Q,S_2}(\phi)
    = -\,\mathbb{E}_{s\sim \mathcal{D}_{\text{mix}},\,a\sim \pi_\phi(\cdot\mid s)}\!\big[Q_\theta(s,a)\big]
    \end{aligned}
    \label{eq:actor_q_s2}
\end{equation}
Here, $\rho$ is the success rate normalization factor. Sampling from $\mathcal{D}_{\text{mix}}$ aligns the actor’s training distribution with the expanded critic support, effectively mitigating covariate shift.

\noindent{\textbf{\textit{S}3. Intervention-Free Online Fine-tuning}}.

In the third stage, we perform online fine-tuning without human intervention. We maintain two replay buffers: an \emph{expert buffer} $\mathcal{D}_\mathrm{E}$ (offline data + successful online trajectories) and a \emph{policy buffer} $\mathcal{D}_\pi$ (all online rollouts). At each update, we sample equally from both to balance retaining competent behaviors and learning from new exploration.

The critic minimizes the standard temporal-difference loss over the union of buffers:
\begin{equation}
    \begin{aligned}
    \mathcal{L}_{Q}^{S_3}(\theta)=&\mathbb{E}_{(s,a,s')\sim(\mathcal{D}_{\text{E}}\cup\mathcal{D}_\pi)}[(Q_{\theta}(s,a)-\mathcal{B}^{\pi}\overline{Q}_{\overline{\theta}}(s,a))^2\Big] \\
    \end{aligned}
    \label{eq:cql_online}
\end{equation}

The actor optimization leverages our Value-Guided Policy Self-Distillation (VGPD) mechanism:
\begin{equation}
\begin{aligned}
    \mathcal{L}_{\pi}^{S_3}(\phi)
    = \lambda\,\mathcal{L}_{\pi}^{Q,S_3}(\phi)
    + \eta\,\mathcal{L}_{\pi}^{Distill,S_3}(\phi),
\end{aligned}
    \label{eq:actor_s3_obj}
\end{equation}
where $\mathcal{L}_{\pi}^{Q,S_3}(\phi)$ encourages high-value actions as in Eq.~\ref{eq:actor_q_s2}. The distillation term $\mathcal{L}_{\pi}^{Distill,S_3}(\phi)$ is detailed below.

\begin{table*}[t]
    \centering
    \small
    \resizebox{\textwidth}{!}{%
    \begin{tabular}{llccccccc}
        \toprule
        \multicolumn{2}{c}{} &
        \multicolumn{6}{c}{Success Rate (\%)} &
        \multirow{2}{*}{Average(\%)} \\
        \cmidrule(lr){3-8}
        Group & Method &
        StackCube & PullCube & PushCube &
        PullCubeTool & PlaceSphere & PickCube & \\
        \midrule
        \multirow{3}{*}{BC}
            % & pi0  & 0 & 14.5 & 70 & 0 & 17 & 18.5 & 20.0 \\
            & Octo\cite{ghosh2024octo} & 0 & 0 & 13 & 0 & 0 & 8.5 & 3.58 \\
            & $\pi_0$\cite{blackP0VisionLanguageActionFlow}  & 60 & 87.5 & 70 & 7.5 &27.5 & 2.5 & 42.5 \\
            & $\pi_{0.5}$\cite{blackP05VisionLanguageActionModel}  & 50 & 87.5 & \textbf{100} & 7.5 & 15 & 5 & 44.17 \\
        \midrule
        \multirow{6}{*}{RL}
            & Cal-QL\cite{nakamoto2023calql}          & 65.2 & 84.9   & 96.7 & 0  & 0 & 24.2   & 45.2 \\
            & CQL\cite{kumar2020cql}             & 0    & 0 & 92.2 & 0  & 0   & 2.1    & 15.7 \\
            & PA-RL\cite{mark2024policy}   & 73.9  & 81.1  & 93.7   & 0   & 0  & 52.4   & 50.2 \\
            & ConRFT (no HIL)\cite{chen2025conrft} & 82.1 & 91.7 & 95.9 & 0  & 69.8 & 87.2   & 71.1 \\
            & \textbf{Ours (Octo)}   & \textbf{94.2} & 99 & \textbf{100} &
                               19 & 85.1 & \textbf{96.7} &
                               82.3 \\
            & \textbf{Ours ($\boldsymbol{\pi}_0$)}   & 93.2 & \textbf{100} & \textbf{100} & \textbf{36.7} & \textbf{97.5} & 94.1 & \textbf{86.9} \\
        \bottomrule
    \end{tabular}%
    }
    \caption{\textbf{Success rates of BC and RL methods on ManiSkill tasks.} We report the final converged success rates across six tasks on three random seeds. For all other reinforcement learning (RL) baselines, we consistently utilize the same offline pre-trained Octo and pi0 model as the backbone for post-training fine-tuning. 
}
    \label{tab:result_overview}
    \vspace{-1.5em}
\end{table*}

\subsection{Value-Guided Policy Self-Distillation (VGPD)}
\label{sec:vgpd}

Our proposed VGPD mechanism can be theoretically grounded as an approximate solution to a \textit{Regularized Policy Improvement} problem with a dynamic data-dependent baseline. Following the formulation in trust-region methods \cite{peng2019awr, nair2020awac}, we aim to maximize the expected return subject to a KL-divergence constraint to prevent the policy from deviating too far from the data support. The objective is defined as:
\begin{equation}
\begin{split}
    \mathcal{J}(\pi) = \mathbb{E}_{s \sim \mathcal{D}} \Big[ & \mathbb{E}_{a \sim \pi(\cdot|s)} [Q_\theta(s, a)] \\
    & - \tau D_{KL}(\pi(\cdot|s) \| \pi_{\text{old}}(\cdot|s)) \Big]
\end{split}
\label{eq:objective}
\end{equation}
where $\pi_{\text{old}}$ serves as the reference proposal distribution. In the context of our online self-distillation, $\pi_{\text{old}}$ corresponds to the current policy $\pi_\phi$. The optimal closed-form solution is the energy-based distribution $\pi^*(a|s) \propto \pi_{\text{old}}(a|s) \exp\left(Q_\theta(s, a)/\tau\right)$.

Since sampling strictly from $\pi^*$ is intractable, we project this optimal policy back into our parameterized policy space $\Pi$ by minimizing $D_{KL}(\pi^* \| \pi_\phi)$, which is equivalent to maximizing the weighted log-likelihood:
\begin{equation}
    \max_\phi \mathbb{E}_{s \sim \mathcal{D}, a \sim \pi_{\text{old}}} \left[ \exp\left(\frac{Q_\theta(s, a)}{\tau}\right) \log \pi_\phi(a|s) \right]
\end{equation}

\noindent\textbf{Filtered Importance Sampling.} VGPD approximates the expectation over $\pi_{\text{old}}$ using a set of $K$ samples $\{\hat{a}_k\}_{k=1}^K \sim \pi_\phi(\cdot|s)$. To mitigate variance from the exponential weights, VGPD introduces a \textit{Dynamic Advantage Filter}. We define a state-dependent baseline as the empirical mean value: $V_{\text{ref}}(s) = \frac{1}{K}\sum_{k=1}^K Q_\theta(s, \hat{a}_k)$.
The update rule applies a mask $\mathbb{I}[Q_\theta(s, \hat{a}_k) \ge V_{\text{ref}}(s)]$, effectively performing a \textit{Positive Advantage Truncation} where samples with negative advantage are discarded.

\noindent\textbf{Proposition 1 (Monotonic Value Improvement).} \textit{Let $\pi_{\text{tgt}}$ be the empirical target distribution constructed by VGPD. Assuming the critic $Q_\theta$ is consistent, the expected Q-value under the target distribution is lower-bounded by $V_{\text{ref}}(s)$, ensuring that the distillation target always represents a monotonic improvement over the average performance of the current policy rollout.}

\noindent\textbf{Implementation.} We implement this theoretical framework as follows. For expert data $\mathcal{D}_\mathrm{E}$, we treat stored actions as optimal, degenerating to behavior cloning:
\begin{equation}
\begin{aligned}
    \mathcal{L}_{\pi}^{Distill,D_E}(\phi)
    = \mathbb{E}_{(s,a)\sim \mathcal{D}_\mathrm{E}}
    \big[-\log \pi_\phi(a\mid s)\big].
\end{aligned}
    \label{eq:vgpd_bc_expert}
\end{equation}

For the policy buffer $\mathcal{D}_\pi$, given a sample $(s, a_{\text{buf}})$, we sample $K$ candidates $\{a_k\}_{k=1}^K$ from $\pi_\phi(\cdot|s)$. We compute the baseline $q_{\text{mean}}(s)$ corresponding to $V_{\text{ref}}(s)$ in Eq.~\ref{eq:q_mean}:
\begin{equation}
    q_{\text{mean}}(s) = \frac{1}{K}\sum_{k=1}^{K} Q_\theta(s,a_k).
    \label{eq:q_mean}
\end{equation}
We define an indicator $\zeta(s) = \mathbf{1}\{q_{\text{mean}}(s) > Q_\theta(s,a_{\text{buf}})\}$. The target distribution is constructed as:
\begin{equation}
\begin{aligned}
    \mu^{\text{VGPD}}(\cdot\mid s)
    =&\;(1-\zeta(s))\,\delta_{a_\mathrm{buf}}(\cdot) \\
    &\;+\zeta(s)\sum_{k=1}^{K} \tilde{w}_k(s)\,\delta_{a_k}(\cdot),
\end{aligned}
    \label{eq:vgpd_target_dist}
\end{equation}
where the weights $\tilde{w}_k(s)$ implement the exponential energy weighting masked by the advantage filter:
\begin{equation}
    \tilde{w}_k(s)
    = \frac{\mathbf{1}\{Q_\theta(s,a_k)\ge q_{\text{mean}}(s)\}
            \exp\big(Q_\theta(s,a_k)/\tau\big)}
            {\sum_{j=1}^{K}\mathbf{1}\{Q_\theta(s,a_j)\ge q_{\text{mean}}(s)\}
            \exp\big(Q_\theta(s,a_j)/\tau\big)}.
    \label{eq:vgpd_weights}
\end{equation}
The final distillation loss is $\mathcal{L}_{\pi}^{Distill,S_3}(\phi) = \mathcal{L}_{\pi}^{Distill,D_E}(\phi) + \mathbb{E}_{\mu^{\text{VGPD}}} [-\log \pi_\phi]$.

%% file: sec/4_experiment.tex
\section{Experiments}
\label{sec:experiment}

\begin{figure}[!ht]
    \centering
    \vspace{-3mm}
    \includegraphics[width=0.9\linewidth]{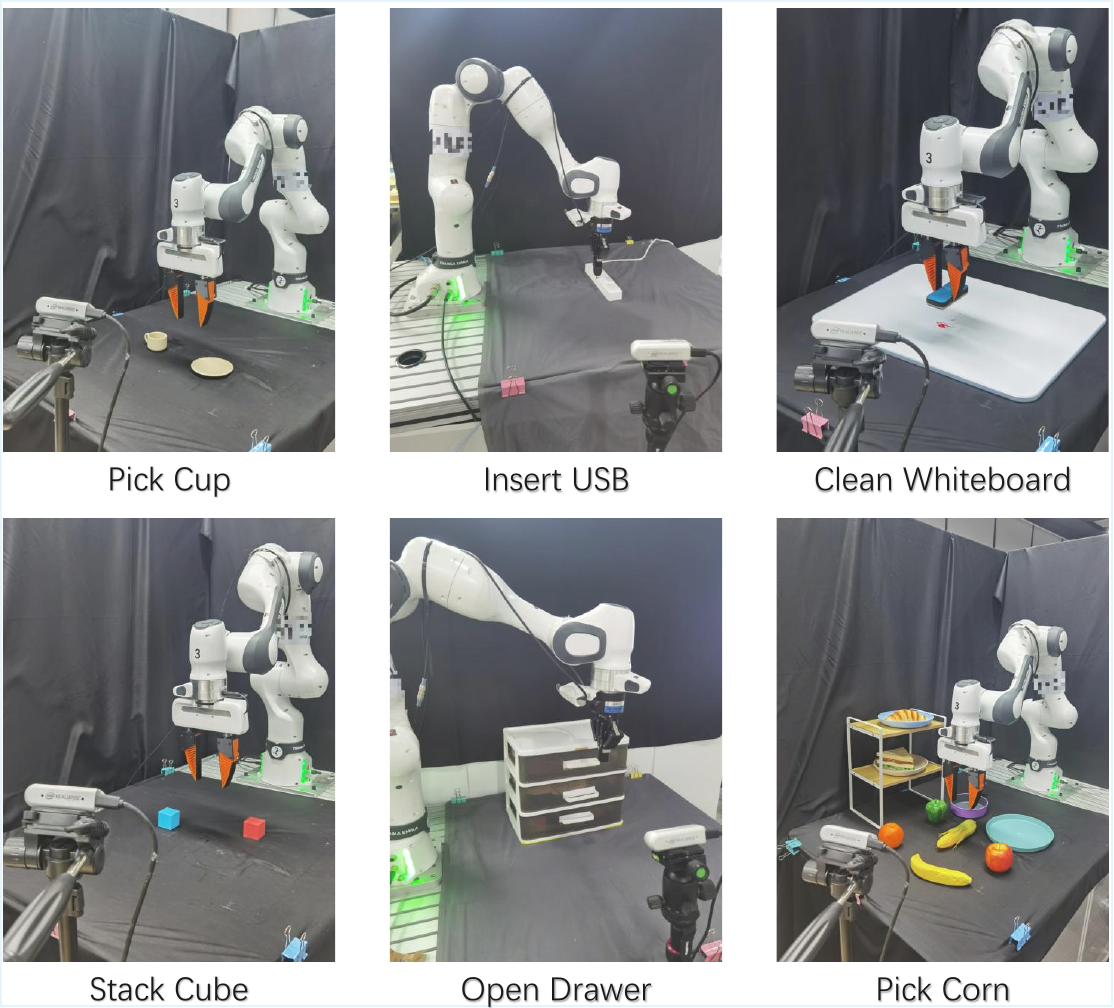}
    \vspace{-2mm}
    \caption{\textbf{Real-world Experiment Tasks}. We conducted real-world experiments using a single-arm Franka robot equipped with two RealSense cameras that supplied complementary visual feedback: a wrist-view and a side-view.}
    \label{fig:experiment_setup}
    % \vspace{-1mm}
\end{figure}

\begin{table}[!ht]
    \centering
    \small
    \setlength{\tabcolsep}{4pt}
    \begin{tabular}{l c c c c}
        \toprule
        \multirow{2}{*}{Task} & \multicolumn{2}{c}{\textbf{Success Rate (\%)}} & \multicolumn{2}{c}{\textbf{Execution Steps}} \\
        \cmidrule(lr){2-3} \cmidrule(lr){4-5}
                              & BC & \textbf{Ours} & BC & \textbf{Ours} \\
        \midrule
        Pick Cup    & 35 & 45\,$\to$\,\textbf{90}  & 134.5 & 125.2\,$\to$\,\textbf{36.4} \\
        Open Drawer & 35 & 55\,$\to$\,\textbf{100} & 57.4  & 63.3\,$\to$\,\textbf{38.2}  \\
        Insert USB  & 75 & 60\,$\to$\,\textbf{100} & 44.5  & 42.7\,$\to$\,\textbf{16.0}  \\
        Pick Corn   & 40 & 45\,$\to$\,\textbf{100} & 142.3 & 153.3\,$\to$\,\textbf{52.4} \\
        Stack Cube  & 40 & 35\,$\to$\,\textbf{100} & 155.2 & 170.4\,$\to$\,\textbf{46.2} \\
        Clean Board & 45 & 50\,$\to$\,\textbf{100} & 142.6 & 136.5\,$\to$\,\textbf{44.3} \\
        \midrule
        \textbf{Avg.} & 45.0 & 48.3\,$\to$\,\textbf{98.3} & 112.8 & 115.2\,$\to$\,\textbf{38.9} \\
        \bottomrule
    \end{tabular}
    \caption{\textbf{Quantitative results on real-world tasks.} Comparison of success rate and execution steps between BC baseline and our FORCE method after online fine-tuning.}
    \label{tab:result_single_col}
    \vspace{-6mm}
\end{table}

\begin{figure*}[!ht]
    \centering
    \vspace{-3mm}
    \includegraphics[width=0.85\textwidth]{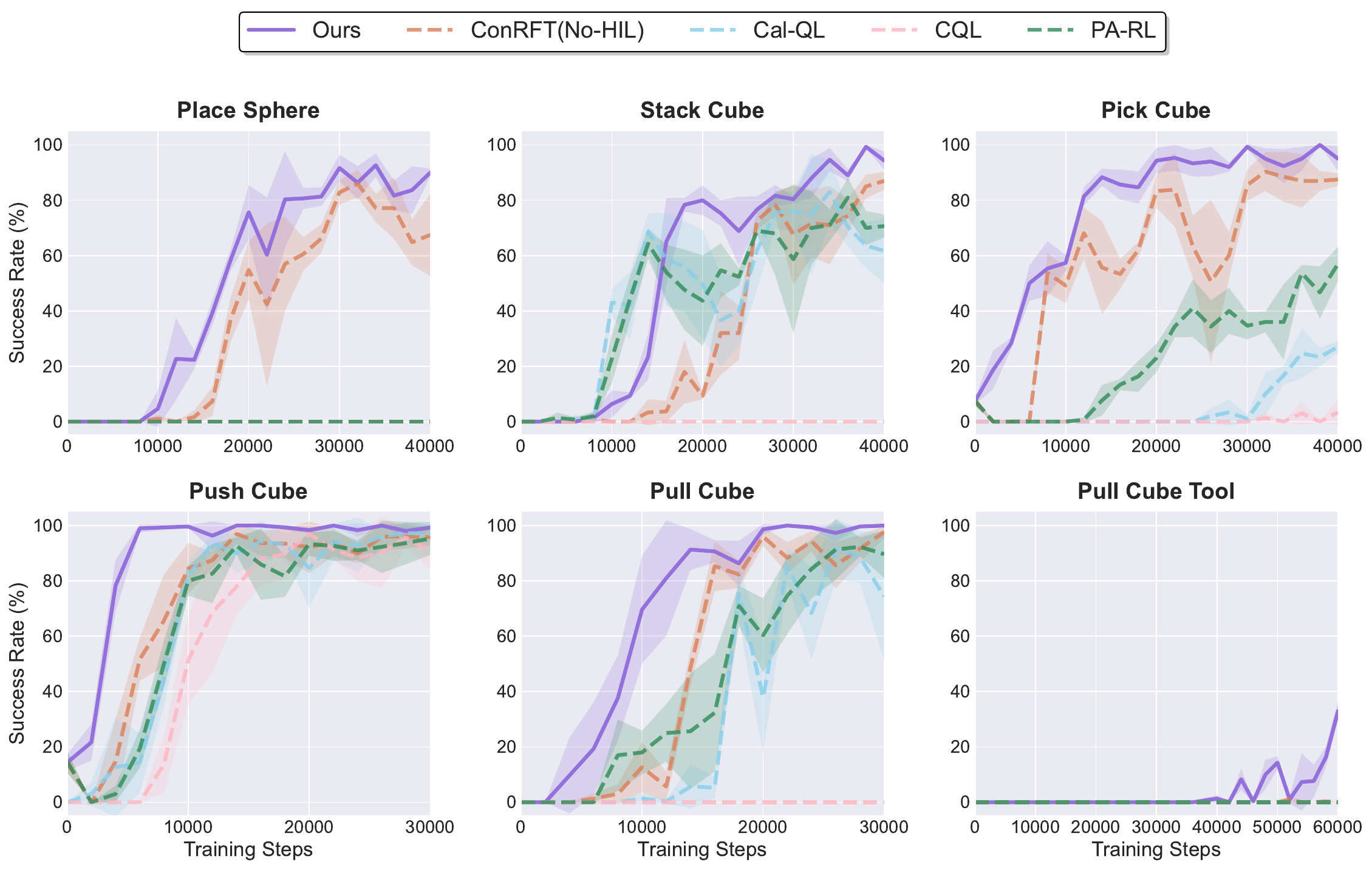}
    \vspace{-2mm}
    \caption{\textbf{Learning Curves.} We present the performance of our proposed FORCE and baselines in ManiSkill tasks. The evaluation spans three random seeds. FORCE consistently demonstrates faster convergence and higher final performance, validating the benefits of distributional warm-up and value-guided updates.}
    \label{fig:training_curve}
    \vspace{5mm}
    \includegraphics[width=0.85\textwidth]{ablation_tasks_training_curves.pdf}
    \vspace{-2mm}
    \caption{\textbf{Ablation Study.} Training curves across four tasks comparing full FORCE against variants without warm-up or VGPD. The results highlight the necessity of mitigating initial distribution shift and filtering policy updates.}
    \label{fig:ablation}
    \vspace{-3mm}
\end{figure*}

\subsection{Overview of Experiments}

We designed our empirical evaluation to verify the theoretical claims of the FORCE framework. Specifically, we investigate whether addressing distributional shift and enforcing regularized updates leads to superior fine-tuning performance. Our analysis focuses on three key questions:

\begin{enumerate}[itemsep=2pt, topsep=4pt, parsep=1pt]
    \item \textbf{Mitigation of Distributional Shift:} Does the Distributional Warm-up stage effectively mitigates the "cold-start" performance collapse typical of offline-to-online transfer?
    \item \textbf{Real-World Reliability:} Does the VGPD mechanism provide a more stable and sample-efficient learning signal compared to standard gradient-based RL updates?
    \item \textbf{Intervention-Free Convergence Validation:} Can FORCE effectively adapt to physical environments and perform consistently in contact-rich tasks without reliance on real-time human correction?
\end{enumerate}
We utilize ManiSkill \cite{taomaniskill3} for large-scale quantitative benchmarking and a physical Franka Emika Panda robot for real-world validation.

% To extensively validate the efficacy and robustness of our proposed three-stage framework, we conducted a comprehensive evaluation across both simulated and real-world environments in this section. Our simulation experiments were conducted on ManiSkill \cite{taomaniskill3}, a widely used benchmark for robotic manipulation that provides diverse simulated environments. To ensure fair comparison, we prioritize this simulated environment for our primary experimental analysis. For real-world validation, we deployed our framework on a Franka Emika Panda robot arm to evaluate its performance in complex physical environments. Aiming for a thorough evaluation of the framework's versatility, we designed our experiments to include heterogeneous manipulation tasks in both domains. The selected tasks span from basic pick-and-place operations to more dynamic actions like pushing and precision-demanding insertion. Finally, we demonstrate the efficacy of our key components through comprehensive ablation studies. Detailed configurations for all experiments can be found in the Appendix.

\subsection{Simulation Experiments}

\textbf{FORCE achieves SOTA performance by effectively expanding policy support.}
Quantitative results in Table~\ref{tab:result_overview} reveal a significant performance gap between FORCE and prior baselines. Our method achieves an average success rate of 82.3\% based on Octo backbone across six tasks, surpassing the strongest baseline, ConRFT, by over 10\%.
A deeper analysis of task-specific failures exposes the limitations of existing methods. Conservative baselines like Cal-QL suffer from severe \textit{policy collapse} on precision-demanding, long-horizon tasks (e.g., \textit{StackCube}: 45.5\%, \textit{PlaceSphere}: 0\%). These methods enforce strict constraints to keep the policy within the support of the offline dataset, which paradoxically hinders the exploration necessary to correct sub-optimal offline behaviors during fine-tuning.
In contrast, FORCE mitigates this "conservative trap" via Distributional Warm-up. By explicitly updating the value function on rollout data \textit{before} releasing the policy for optimization, FORCE effectively expands the valid support of the critic. This pre-conditioning reduces the \textit{extrapolation error} at the boundaries of the dataset, allowing the agent to explore the state space safely and mastering complex tasks like \textit{PullCube} (100\%) and \textit{PickCube} (96.4\%) where baselines stall.

\textbf{VGPD enables rapid convergence and stability.}
The learning curves in Fig.~\ref{fig:training_curve} offer a granular view of the optimization dynamics. FORCE is characterized by a steep, monotonic ascent, contrasting sharply with the erratic performance profiles of standard RL fine-tuning.
On the \textit{PushCube} task, FORCE converges to 100\% success within just 5,000 steps—achieving a 32\% reduction in sample complexity compared to ConRFT.
More critically, the curves highlight the stabilizing role of VGPD. While baselines frequently exhibit "performance chattering"—sharp drops caused by \textit{value overestimation} and noisy gradients guiding the policy into distinct failure modes—FORCE maintains robust stability.
This validates that the Regularized Policy Improvement objective (Eq.~\ref{eq:objective}) acts as an effective \textit{dynamic filter}. By weighting updates based on the value advantage, VGPD significantly improves the signal-to-noise ratio of the learning gradient. It ensures that the policy is updated only towards transitions with high verified value, thereby preventing the "unlearning" of useful primitives and facilitating asymptotic optimality.

\textbf{FORCE demonstrates remarkable compatibility with various VLA backbones.} Experimental results demonstrate that FORCE not only significantly enhances the performance of weaker backbones, such as Octo, but also fully realizes the potential of high-capacity foundations like $\pi_0$. This is particularly evident in high-difficulty tasks, like PullCubeTool task, where the $\pi_0$-based implementation achieved results vastly outperforming all other comparative methods, thereby underscoring the immense potential of FORCE as a general VLA post-training framework.

\subsection{Ablation Study: Dissecting the Mechanism}

To isolate the contributions of Distributional Warm-up and VGPD, we conducted an ablation study summarized in Table~\ref{tab:steps80} and Fig.~\ref{fig:ablation}. We introduce the metric \textbf{Steps@80\%}—the minimum environment steps required to reach 80\% success—to rigorously quantify sample efficiency.

\textbf{Distributional Warm-up prevents cold-start collapse.}
Comparing the "w/o Pre-Cal" variant (Orange) to the full method (Purple) in Fig.~\ref{fig:ablation} reveals the critical role of the warm-up stage. Without warm-up, the policy suffers from immediate performance degradation at the onset of online training. This confirms that the initial covariate shift between the offline dataset and the online policy is the primary driver of instability. By expanding the Q-function support via on-policy rollouts \textit{before} full fine-tuning, FORCE ensures the critic is well-defined for the initial exploration steps.

\textbf{VGPD ensures asymptotic optimality.}
The removal of VGPD (Green) results in a premature performance plateau. As shown in Table~\ref{tab:steps80}, the "w/o VGPD" variant requires significantly more steps to reach competence (e.g., 20k vs 12k for PickCube) or fails to reach the 80\% threshold entirely within the budget. This indicates that standard Q-guided updates are insufficient for fine-grained policy refinement in the later stages of training. VGPD's dynamic advantage filtering effectively removes noise from the gradient, allowing the policy to converge to a higher-performing solution. Overall, FORCE reduces the average Steps@80\% by 32.5\%, demonstrating that both components are essential for maximizing sample efficiency.

\begin{table}[h]
\centering
\small
\begin{tabular}{lccc}
\toprule
Task & w/o Pre-Cal & w/o Pre-Cal\&VGPD & \textbf{Ours} \\
\midrule
StackCube    & 28k & 28k & \textbf{18k} \\
PickCube     & 14k & 20k & \textbf{12k} \\
PushCube     &  8k & 10k & \textbf{4k}  \\
PlaceSphere  & 30k & 30k & \textbf{20k} \\
\bottomrule
\end{tabular}
\caption{\textbf{Steps@80\% success rate.} Lower implies higher sample efficiency. FORCE achieves the target threshold significantly faster than ablated variants.}
\label{tab:steps80}
\vspace{-1em}
\end{table}

\subsection{Adaptive Distillation Dynamics} \label{sec:sim_analysis}

\begin{figure}[h]
  \centering
  \includegraphics[width=0.5\textwidth]{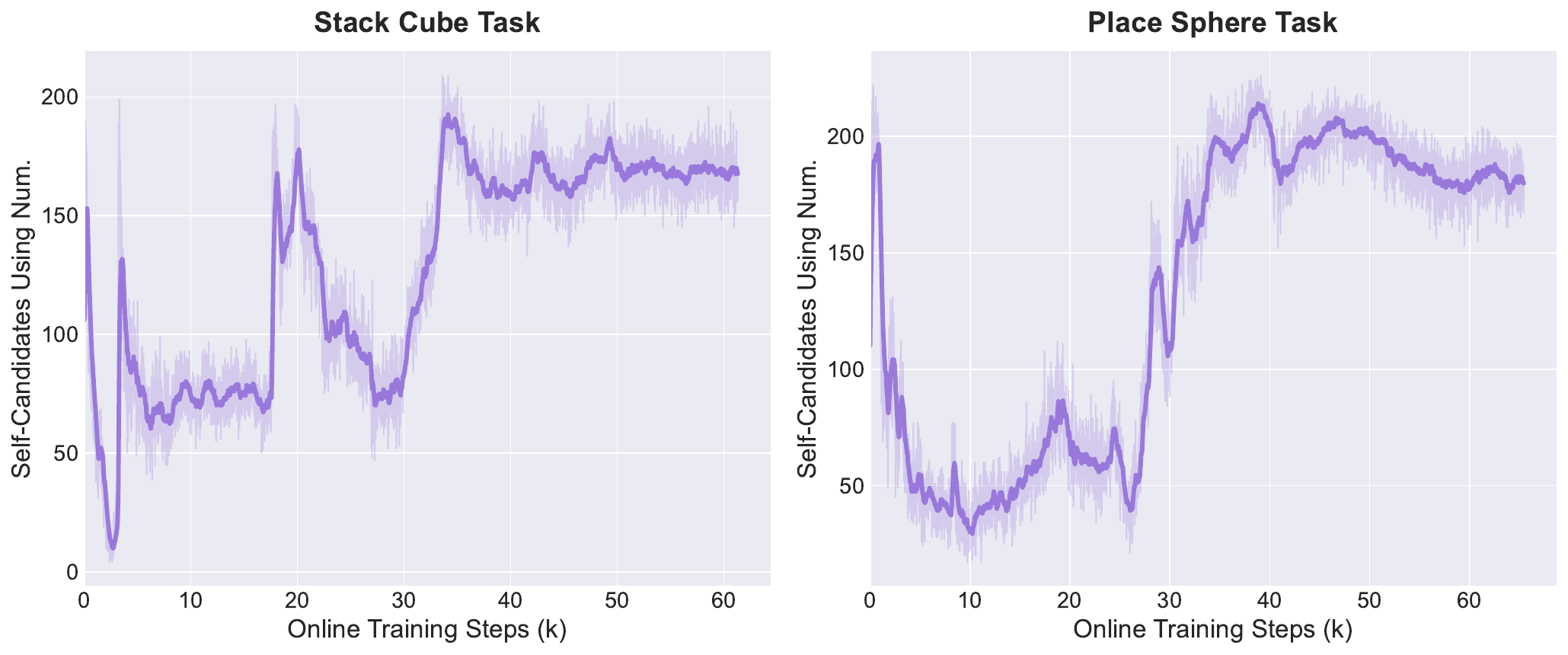}
  \caption{\textbf{Adaptive Nature of VGPD.} We visualize the proportion of "active" learning targets derived from the policy itself (Self-Distillation) versus the offline buffer over time. The mechanism naturally transitions from cloning to exploration.}
  \label{fig:candidate}
  \vspace{-5mm}
\end{figure}

To understand how VGPD regulates the learning process, we analyze the source of the distillation targets throughout training (Fig.~\ref{fig:candidate}). Specifically, we track the ratio of targets derived from successful self-generated rollouts (on-policy) versus the static expert buffer (off-policy).

\textbf{VGPD realizes an automatic curriculum.}
The analysis reveals a dynamic transition in data utilization. In the early training phase, the critic assigns lower values to the immature policy's proposals compared to the expert demonstrations. Consequently, the dynamic advantage filter (Eq.~\ref{eq:vgpd_weights}) heavily weights the expert actions, acting effectively as a stabilizer akin to Behavior Cloning.
% \textbf{Notably, the curves often exhibit pronounced fluctuations at the beginning of online training,} which is consistent with the policy first encountering OOD states and producing unstable rollouts; in this regime, the critic-policy mismatch leads to volatile acceptance of self-candidates.
As the policy improves, the proportion of self-generated actions passing the advantage filter increases. In the later stages (e.g., on PushCube), the algorithm predominantly distills from on-policy data, allowing for fine-grained optimization beyond the sub-optimal demonstrations.
This confirms that VGPD does not require manual tuning of the exploration-exploitation balance; instead, the value-guided mechanism naturally implements an adaptive curriculum—transitioning from \textbf{conservative imitation} to \textbf{aggressive self-improvement} based on the policy's real-time competence.

\subsection{Real-World Experiments}

\textbf{Reliability under Physical Constraints via Value-Calibrated Warm-Up.}
As shown in Table~\ref{tab:result_single_col}, the BC baseline struggles with an average success rate of 45.0\%, frequently failing in precision-dependent tasks like \textit{Insert USB} (75\%) and \textit{Open Drawer} (35\%). In the physical world, slight calibration errors or sensor noise can push the state distribution out of the BC policy's support, leading to compounding errors.
FORCE dramatically boosts the average success rate to \textbf{98.3\%}, achieving 100\% on 4 out of 6 tasks.
We attribute this robustness to the \textit{Distributional Warm-up} stage. By adapting the value function's support to the online policy distribution \textit{before} aggressive updates, FORCE mitigates the risk of "policy collapse" during the initial online phase. This is particularly critical in real-world settings where undefined value estimates can lead to dangerous, erratic robot motions. FORCE effectively maintains a safe exploration corridor, allowing the agent to recover from OOD states induced by physical noise.

\textbf{Execution Efficiency and Motion Quality via Value Guidance.}
Beyond binary success metrics, execution efficiency serves as a proxy for motion quality. Table~\ref{tab:result_single_col} reveals a substantial reduction in execution steps, dropping from 112.8 (BC) to \textbf{38.9} (FORCE).
Qualitatively, BC policies on real hardware often exhibit "hesitant" behaviors—jittery motions or localized loops—when encountering states marginally different from the training set.
In contrast, FORCE reduces the steps required for \textit{Stack Cube} by nearly $4\times$ (170.4 $\to$ 46.2) and \textit{Clean Board} by $3\times$.
This efficiency validates the \textit{VGPD} mechanism in the physical domain. By rigorously filtering gradients based on value advantage, VGPD suppresses the high-variance, suboptimal actions often generated during stochastic exploration. 

%% file: sec/5_conclusion.tex
\section{Conclusion}
\label{sec:conclusion}
In this work, we introduced FORCE, a principled framework for VLA fine-tuning that explicitly addresses the distributional mismatches inherent in offline-to-online transfer. By expanding the Q-function support via a Distributional Warm-up and enforcing monotonic improvement through Value-Guided Policy Distillation, FORCE effectively mitigates the catastrophic unlearning and high variance typical of standard RL fine-tuning. Our empirical results across simulation and real-world domains confirm that FORCE achieves state-of-the-art performance and superior sample efficiency without requiring human intervention, enabling robust policy adaptation in contact-rich environments.

Despite these promising results, FORCE entails a computational overhead due to the requirement of sampling multiple action candidates ($K$) during the VGPD step. This inference cost can be significant when fine-tuning VLA backbones. Future work could address this bottleneck by investigating action caching mechanisms or utilizing off-policy samples from previous iterations to approximate the distillation target. Additionally, while our current implementation utilizes task-specific critics, extending this framework to leverage Generalist Reward Models represents a vital direction for scaling intervention-free fine-tuning to open-ended, long-horizon tasks.

\section*{Impact Statement}

This work develops an RL fine-tuning framework for robotic Vision-Language-Action policies to improve sample efficiency and stability in offline-to-online adaptation. It may reduce reliance on human intervention and accelerate deployment in real-world manipulation. Potential risks include unsafe behaviors under distribution shift or uncontrolled exploration; therefore, practical use should follow standard safety constraints and thorough evaluation in controlled settings.

% In this paper, to address the critical instability and inefficiency of offline-to-online VLA fine-tuning, we introduce a novel three-stage framework named FORCE. Our framework stabilizes the Q-function via a novel pre-online phase and then employs value-guided self-distillation to purify both online exploration and offline data streams. Our extensive real-world and simulation experiments demonstrate that FORCE achieves state-of-the-art performance, yielding a 79\% absolute improvement in success rates and outperforming prior RL methods by 10\%, accelerates training by 32.5\%. Critically, it eliminates the common success rate drop, and achieves this robust performance without any human intervention, marking a significant step towards deploying capable and autonomous robotic agents.

% While our current framework demonstrates robust performance with sparse rewards, its sample efficiency could be further accelerated. To this end, one could investigate integrating a pre-trained General Reward Model to serve as a more informative critic. Such a model could provide more granular, step-wise feedback, potentially guiding the policy to master even more complex and long-horizon tasks, further bridging the gap toward intervention-free robotic systems.

% reduces the steps required to reach an 80% success rate by 67%,

%% file: sec/X_suppl.tex
% Appendix content (ICML): main.tex will switch to \appendix before inputting this file.

\section{Algorithm Illustration} \label{sec:algorithm}
The whole pipeline of our method is outlined in Algorithm \ref{alg:pieline}.

\begin{algorithm}[h]
    \caption{Training Pipeline}
    \label{alg:pieline}
    \scalebox{1}{
    \begin{minipage}{1.2\columnwidth}
    \begin{algorithmic}[1]
        \REQUIRE Pretrained policy $\pi_{\text{pre}}$, expert dataset $\mathcal{D}_E$, environment $\mathcal{M}$
        \ENSURE Fine-tuned policy $\pi_{\phi}$
        \STATE Initialize $\pi_{\phi} \gets \pi_{\text{pre}}$, critic $Q_{\theta}$
        \STATE \textcolor{gray!90}{\# Stage I: Offline reinforcement fine-tuning}
        \FOR{offline training iterations}
            \STATE Sample $(s,a,r,s')$ from $\mathcal{D}_E$
            \STATE Update critic $Q_{\theta}$ and policy $\pi_{\phi}$ using Equation \ref{eq:calql_offline} and Equation \ref{eq:offline_actor_loss}
        \ENDFOR
        \STATE \textcolor{gray!90}{\# Stage II: Value pre-calibration}
        \STATE Collect on-policy rollouts $\mathcal{D}_R$ by executing $\pi_{\phi}$ in $\mathcal{M}$
        \STATE $\mathcal{D}_{\text{mix}} \gets \mathcal{D}_E \cup \mathcal{D}_R$ with success labels $y$
        \FOR{stabilization iterations}
            \STATE Sample $(s,a,r,s',y)$ from $\mathcal{D}_{\text{mix}}$
            \STATE Update critic $Q_{\theta}$ and policy $\pi_{\phi}$ using Equation \ref{eq:calql_offline} and Equation \ref{eq:actor_s2_obj}
        \ENDFOR
        \STATE \textcolor{gray!90}{\# Stage III: Online reinforcement fine-tuning with VGPD}
        \STATE Initialize replay buffers $\mathcal{D}_E$ (expert) and $\mathcal{D}_\pi$ (policy)
        \WHILE{online training}
            \STATE Sample equal-size batches from $\mathcal{D}_E$ and $\mathcal{D}_\pi$
            \FOR{each $(s,a)$ in batch}
                \IF{$(s,a)$ from $\mathcal{D}_E$}
                    \STATE $a^* \gets a$ \COMMENT{Expert action as target}
                \ELSE
                    \STATE $q_{\text{buf}} \gets Q_{\theta}(s,a)$
                    \STATE Sample $K$ actions $\{a_k\}$ from $\pi_{\phi}(s)$
                    \STATE Compute $q_k = Q_{\theta}(s,a_k)$ for $k=1,\dots,K$
                    \STATE $q_{\text{mean}} \gets \frac{1}{K}\sum_{k=1}^K q_k$
                    \IF{$q_{\text{buf}} \ge q_{\text{mean}}$}
                        \STATE $a^* \gets a$ \COMMENT{Use buffered action}
                    \ELSE
                        \STATE Select actions $\{a_k: q_k \ge q_{\text{mean}}\}$
                        \STATE Compute weights $\tilde{w}_k$ via softmax of $q_k/\tau$
                        \STATE Distillation targets: actions $\{a_k\}$ with weights $\{\tilde{w}_k\}$
                    \ENDIF
                \ENDIF
            \ENDFOR
            \STATE Update critic $Q_{\theta}$ and policy $\pi_{\phi}$ using Equation \ref{eq:cql_online} and Equation \ref{eq:actor_s3_obj}
        \ENDWHILE
    \end{algorithmic}
    \end{minipage}
    }
\end{algorithm}

\section{Details About Experiments} \label{sec:real_world}
\subsection{Experiment Setup}
We deploy FORCE on a real-world Franka Emika Panda robot equipped with two Intel RealSense D435 cameras, capturing wrist and side views. The end-effector is either a Robotiq gripper or a soft gripper adapted from UMI\cite{chi2024universal}, depending on the task. The model input consists of two RGB images and a 7-dimensional proprioceptive state vector, comprising the 6-dimensional delta end-effector pose and the 1-dimensional gripper status. All other configurations remain consistent with the simulation experiments. Both data collection and policy execution operate at a control frequency of 10 Hz. Task success is determined by human operators.
We carried out six different real-robot experimental tasks, with detailed setups described in Sec. \ref{sec:task-description}. For our real-world experiments, the setup environment are shown in Fig. \ref{fig:experiment_setup}.

\begin{figure*}[ht]
  \centering
  \includegraphics[width=0.9\textwidth]{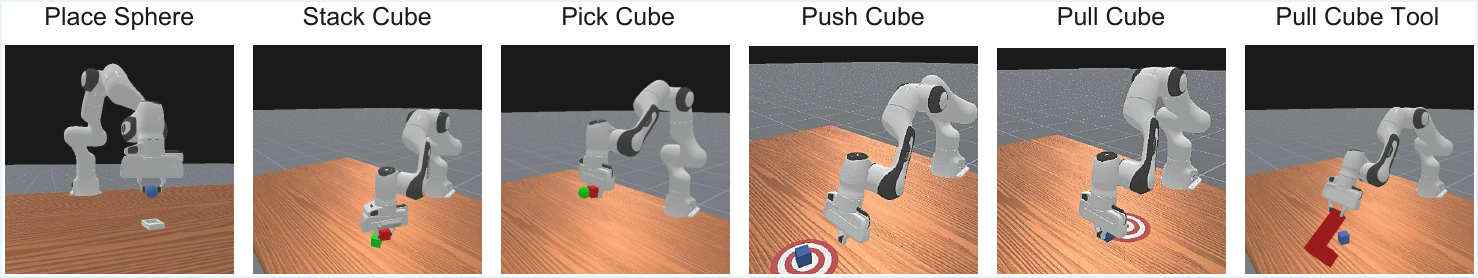}
  \caption{\textbf{Simulation experiments on six ManiSkill manipulation tasks.} 
    \textit{PlaceSphere}: place a small sphere onto a goal pad; 
    \textit{StackCube}: stack one cube on another; 
    \textit{PickCube}: pick and lift a target cube; 
    \textit{PushCube}: push a cube into a goal region; 
    \textit{PullCube}: pull a cube into a goal region; 
    \textit{PullCubeTool}: use an L-shaped tool to drag a cube that starts outside the robot’s reachable workspace into the goal region.}
  \label{fig:sim_exp}
  \vspace{-5mm}
\end{figure*}

\begin{figure*}[ht]
    \centering
    \includegraphics[width=0.77\textwidth]{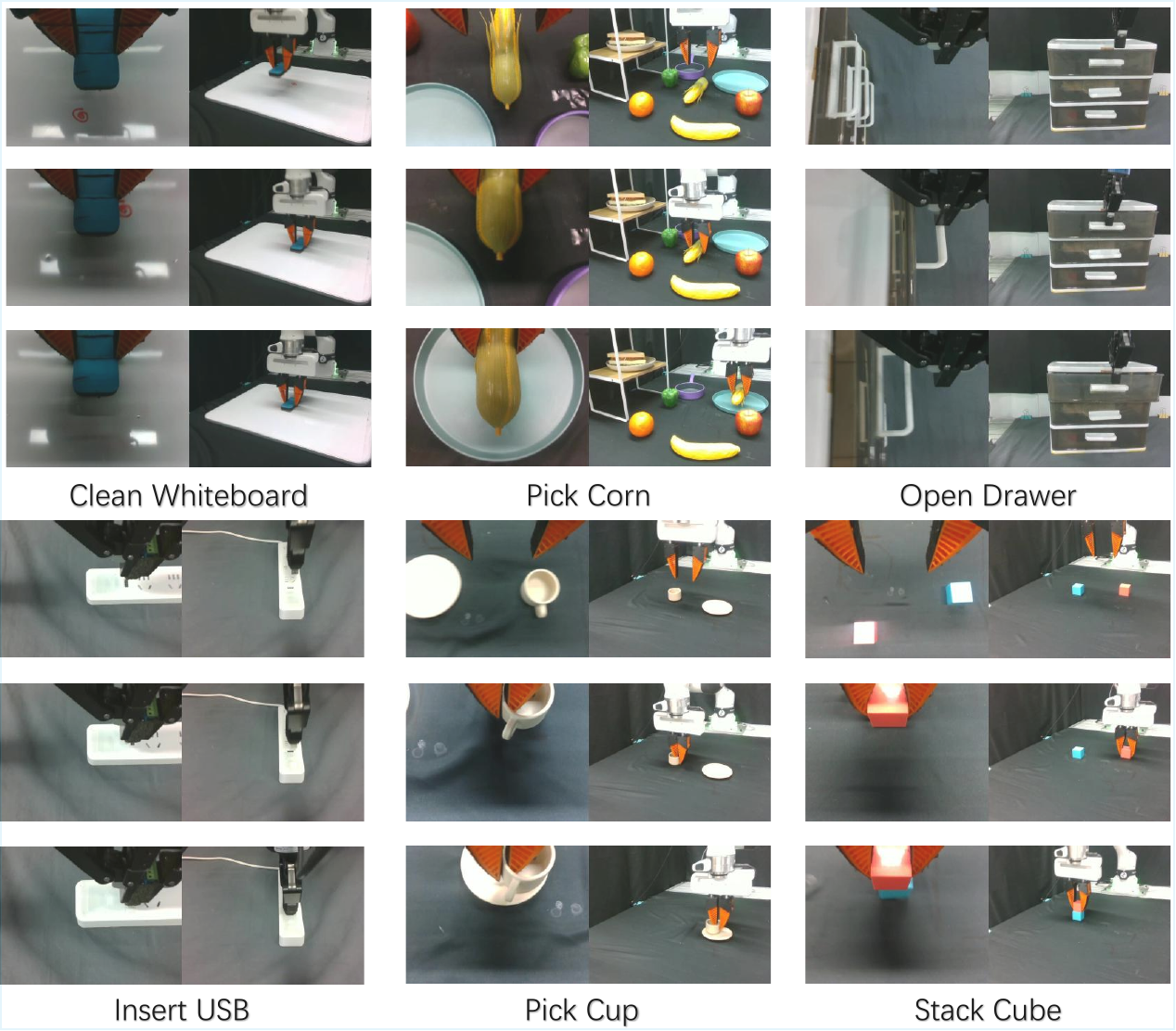}
    \caption{Camera view of successful trajectories on real-world experiment tasks.}
    \label{fig:real_trajectories}
    \vspace{-5mm}
\end{figure*}

\subsection{Task Descriptions} \label{sec:task-description}

\textbf{Pick Cup:} Pick up a cup and place it onto a plate.
\textbf{Open Drawer:} Open the top drawer of a cabinet.
\textbf{Insert USB:} Insert a USB into a port with precise alignment.
\textbf{Pick Corn:} Pick up a corn from a cluttered environment and place it on a plate.
\textbf{Stack Cube:} Stack a red cube on top of a blue cube.
\textbf{Clean Whiteboard:} Erase doodles from a whiteboard.

The training setups for all six tasks are shown in Table \ref{tab:all_tasks_training}. All tasks share common hyperparameters: 50 offline demonstrations, $(\alpha, \beta, \eta) = (0.01, 1.0, 0.1)$ for offline fine-tuning, and $(\beta, \eta) = (0.5, 1.0)$ for online fine-tuning.

\subsection{Implementation Details}

\textbf{Network Architecture.} Our method employs a diffusion-based consistency policy with a 2-layer MLP architecture (hidden size 256) using Tanh activations. The diffusion process operates with noise scheduling parameters $\sigma_{\min}=0.002$, $\sigma_{\max}=80.0$, $\sigma_{\text{data}}=0.5$, and follows a power schedule with $\rho=7.0$ across $M=40$ discretization steps. The noise levels are computed as $\sigma_i = (\sigma_{\min}^{1/\rho} + \frac{i-1}{M-1}(\sigma_{\max}^{1/\rho} - \sigma_{\min}^{1/\rho}))^\rho$ for $i=1,\ldots,M$. For Q-function estimation, we utilize a dual-critic ensemble architecture where each critic is a 2-layer MLP with 256 hidden units and Tanh activations. Visual observations from two RGB cameras ($640\times480$ resolution) are encoded through a pretrained ResNet-18 backbone, with temporal stacking over 2 consecutive frames.

\begin{table*}[h]
    \centering
    \caption{Task-specific training parameters for real-world experiments.}
    \label{tab:all_tasks_training}
    \begin{tabular*}{\textwidth}{@{\extracolsep{\fill}} l cccccc @{}}
        \toprule
        \textbf{Parameter} & \textbf{Pick Cup} & \textbf{Open Drawer} & \textbf{Insert USB} & \textbf{Pick Corn} & \textbf{Stack Cube} & \textbf{Clean Board} \\
        \midrule
        Max episode length & 500 & 300 & 300 & 300 & 1000 & 1000 \\
        Offline steps & 20k & 24k & 20k & 14k & 26k & 20k \\
        Transition steps & 24k & 26k & 24k & 16k & 30k & 24k \\
        Online steps & 32k & 30k & 28k & 20k & 36k & 32k \\
        Negative reward & $-0.05$ & $-0.01$ & $-0.01$ & $-0.05$ & $-0.01$ & $-0.01$ \\
        Online training time & 40 min & 20 min & 20 min & 20 min & 30 min & 40 min \\
        \bottomrule
    \end{tabular*}
    \vspace{-3mm}
\end{table*}

\textbf{Hyperparameter.} We train the agent using Adam optimizer with learning rate $3\times10^{-4}$ for both policy and critic networks. Training proceeds with batch size 256, discount factor $\gamma=0.99$, and exponential moving average updates for target networks with coefficient $\tau=0.005$. 

Our training pipeline consists of three progressive phases with adaptive loss weights: (1) an \textit{offline pretraining phase} using demonstration data with loss weights $\lambda_Q=0.1$ and $\lambda_{BC}=1.0$ to prioritize imitation learning; (2) a \textit{transition phase} where we gradually increase the Q weight with $\lambda_Q=0.3$ and $\lambda_{BC}=1.0$; (3) an \textit{online fine-tuning phase} with $\lambda_Q=1.0$ and $\lambda_{BC}=0.5$ to favor policy improvement over data imitation. 

\textbf{Self-Distillation and Regularization.} To enable adaptive behavior cloning, we sample $N=20$ action candidates per state and select learning targets based on Q-value comparison with a margin threshold $\delta=0.1$. The policy loss combines behavior cloning and Q-maximization objectives. For conservative regularization, we apply CQL with coefficient $\alpha=0.1$, evaluating 10 uniformly sampled out-of-distribution actions per training batch. The critic is updated twice as frequently as the policy (2:1 ratio) to ensure stable value estimation. Network parameters are synchronized to actors every 50 gradient steps during distributed training.

\textbf{Reward Design.} The reward function provides sparse completion signals: $+10$ for successful task execution, $(-0.01,-0.05)$ per timestep to encourage efficiency, and $-0.2$ for each gripper actuation to minimize unnecessary movements.